\documentclass[12pt]{l4dc2021} 

\usepackage{amsmath}
\usepackage{amsfonts}
\usepackage{amssymb}
\usepackage{xcolor}
\usepackage{wrapfig}
\usepackage{algorithm}
\usepackage{algorithmic}

\usepackage{url}
\usepackage[ruled]{algorithm2e}

\usepackage{hyperref}

\definecolor{darkcerulean}{rgb}{0.03, 0.27, 0.49}
\definecolor{cornellred}{rgb}{0.7, 0.11, 0.11}

\makeatletter
\DeclareRobustCommand{\cev}[1]{%
  {\mathpalette\do@cev{#1}}%
}
\newcommand{\do@cev}[2]{%
  \vbox{\offinterlineskip
    \sbox\z@{$\m@th#1 x$}%
    \ialign{##\cr
      \hidewidth\reflectbox{$\m@th#1\vec{}\mkern4mu$}\hidewidth\cr
      \noalign{\kern-\ht\z@}
      $\m@th#1#2$\cr
    }%
  }%
}
\makeatother

\newcommand{\adjoint}{\alpha}

\newtheorem*{claim*}{Claim}

\title[Continuous-Time Policy Gradients]{Faster Policy Learning with Continuous-Time Gradients}
\usepackage{times}

\author{\Name{Samuel Ainsworth} \Email{skainswo@cs.washington.edu}\\
 \Name{Kendall Lowrey} \Email{klowrey@cs.washington.edu}\\
 \Name{John Thickstun} \Email{thickstn@cs.washington.edu}\\
 \Name{Zaid Harchaoui} \Email{zaid@cs.washington.edu}\\
 \Name{Siddhartha Srinivasa} \Email{siddh@cs.washington.edu}\\
 \addr Paul G. Allen School of Computer Science and Engineering, University of Washington
}

\begin{document}

\maketitle

\vspace{-11mm}
\begin{abstract}%
We study the estimation of policy gradients for continuous-time systems with known dynamics. By reframing policy learning in continuous-time, we show that it is possible to construct a more efficient and accurate gradient estimator. 
The standard back-propagation through time estimator (BPTT) computes exact gradients for a crude discretization of the continuous-time system. In contrast, we approximate continuous-time gradients in the original system.
With the explicit goal of estimating continuous-time gradients, we are able to discretize adaptively and construct a more efficient policy gradient estimator which we call the Continuous-Time Policy Gradient (CTPG).
We show that replacing BPTT policy gradients with more efficient CTPG estimates results in faster and more robust learning in a variety of control tasks and simulators.
\end{abstract}


\section{Introduction}\label{sec:intro}


Many robotic control problems can be reduced to finding a desirable trajectory through a physical system governed by piece-wise smooth dynamics, e.g. control of ground vehicles, robot arms, quadrotors, and so forth. When the dynamics of a system are known we can efficiently optimize a trajectory via first-order methods, constructed using explicit gradients of the dynamics. We can use these gradients to solve open-loop trajectory optimization problems~\citep{trajopt_legged_robots,8358969,8403260} or to train a closed-loop feedback controller, i.e. a policy~\citep{8750823,8594312,8593843}. In this paper we focus on the latter problem of policy learning, although our insights can be applied equally well to trajectory optimization.

Computing exact policy gradients for most continuous-time dynamical systems proves intractable. Policy gradient algorithms rely on discrete numerical methods to estimate a robot's trajectory under a policy, and the sensitivity of this trajectory to the robot's actions dictated by the policy. The standard numerical estimate of policy gradients is back-propagation through time (BPTT); this algorithm abandons study of the continuous problem and immediately discretizes time, computing exact policy gradients with respect to discretized dynamics. The discretization step-size controls the accuracy of BPTT, resulting in a tradeoff between the accuracy of the gradient estimates and the computational cost of computing an estimate.

In this paper we defer discretization and begin by characterizing the continuous-time policy gradient. Rather than computing an exact gradient of a discretized system, we instead compute an approximate gradient of the continuous system. When we discretize to compute this approximation, we do so with the explicit goal of constructing an efficient estimate of the continuous-time policy gradient. This approach takes inspiration from the numerical ODE literature, as well as the neural ODE \citep{chen2018neural}. We are motivated by the following questions:

\textbf{Q1: Can we compute more accurate estimates of the policy gradient?} BPTT introduces numerical error by discretizing the robot trajectory under a policy. This error injects variance into the learning process, which could slow down learning. By computing a more accurate estimate of the policy gradient, can we optimize a policy in fewer iterations?

\textbf{Q2: Can we compute policy gradient estimates more efficiently?} Suppose we want a target level of accuracy for our policy gradient estimates. We can achieve the target accuracy by controlling the discretization step-size for BPTT. By using better numerical algorithms, can we achieve the same accuracy with a smaller computational budget?


We give affirmative answers to both \textbf{Q1} and \textbf{Q2}. We propose a new policy gradient estimator (Section \ref{sec:algos}) which we call the Continuous-Time Policy Gradient (CTPG). This estimator strictly improves upon the BPTT estimator, in the sense that the tradeoff curve between gradient accuracy and computational budget for CTPG Pareto-dominates the tradeoff curve for BPTT. In the spirit of algorithms like SGD, we ask: is it possible to sacrifice accuracy of the gradient estimates in order to optimize more efficiently? We find that a certain amount of numerical accuracy is required of our policy gradient estimator in order for optimization to succeed: unlike SGD, poor numerical estimates of policy gradients are not unbiased. However, in a variety of experimental settings (Section \ref{sec:exp}) we demonstrate that replacing expensive BPTT estimates with relatively inexpensive CTPG estimates of comparable accuracy results in faster overall learning.

\section{Learning with Policy Gradients}\label{sec:setting}

Our goal is to learn how to act within a deterministic physical system while minimizing a cost functional. Given a model of the dynamics $f : \mathbb{R}^d \times \mathbb{R}^k \to \mathbb{R}^d$ and time-varying control inputs (actions) $u : [0,\infty) \to \mathbb{R}^k$, the state of the system $x : [0,\infty) \to \mathbb{R}^d$ is governed by a first-order ordinary differential equation
\begin{equation}\label{eqn:dynamics}
\frac{dx(t)}{dt} = f(x(t), u(t)).
\end{equation}
Guided by a local cost (reward function) $w : \mathbb{R}^d \times \mathbb{R}^k \to \mathbb{R}$ of taking action $u$ in state $x$, and a terminal cost $J:\mathbb{R}^d \to \mathbb{R}$, we seek controls $u$ that minimize the global cost over a trajectory of length $T$:
\begin{equation}
\label{eq:trajopt}
\begin{aligned}
& \underset{u(t)}{\text{minimize}} & & \int_0^T w(x(t), u(t)) \,dt + J(x(T)) \\
& \text{subject to} & & \frac{dx(t)}{dt} = f(x(t), u(t)). \\
\end{aligned}
\end{equation}

We are interested in learning feedback controllers in the form of policies $u(t) = \pi_\theta (x(t))$ with parameters $\theta$. For any fixed initial conditions $x(0) = x_0$, the optimization Eq. \eqref{eq:trajopt} is a trajectory optimization problem with parameters $\theta$. For any fixed policy $\pi_\theta(x(t))$, computation of the trajectory $x(\cdot)$ is an initial value problem that is completely determined by the starting state $x_0$. The global loss of following the policy $\pi_\theta$ along a trajectory $x(\cdot)$ is given by
\begin{equation}\label{eqn:trajectoryloss}
\mathcal{L}(x(0), \theta) \triangleq \int_0^T w(x(t), \pi_\theta(x(t))) \,dt + J(x(T)).
\end{equation}
We seek to learn a policy that generalizes over a distribution of initial states $x_0 \sim \rho_0$, motivating us to minimize the expected loss $\mathcal{L}(\theta) \triangleq \mathop{\mathbb{E}}_{x_0\sim\rho_0}\mathcal{L}(x_0,\theta)$.

\begin{figure}
    \centering
    \includegraphics[width=0.9\textwidth]{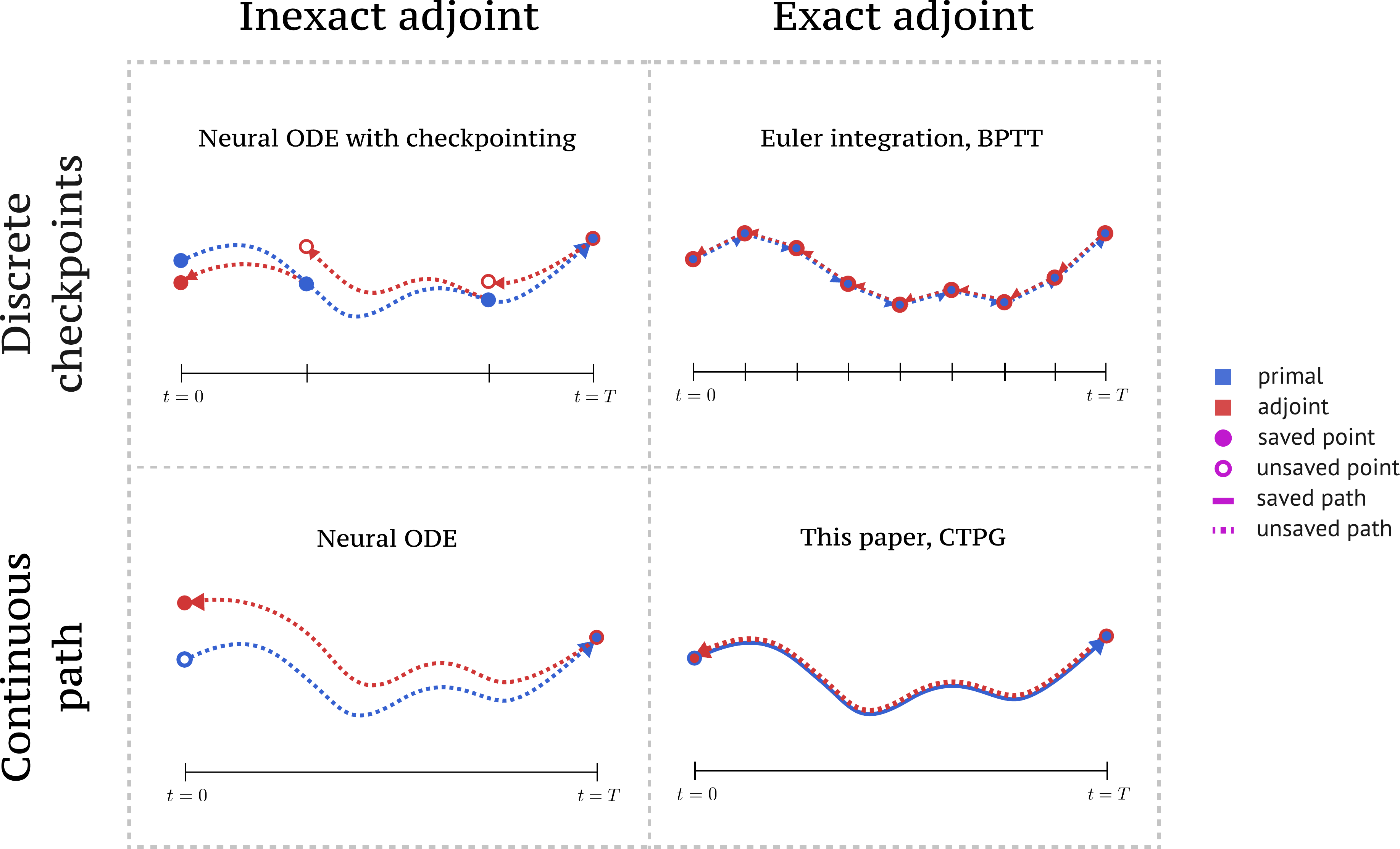}
    \caption{Comparing policy gradient estimators. Neural ODE (Sec.~\ref{sec:neuralode}) approximates continuous trajectories, but accumulated error in its adjoint estimates. Neural ODE with checkpointing mitigates this error, but still accumulates error between checkpoints. Euler integration with BPTT (Sec.~\ref{sec:bptt}) stores a discretized trajectory, which prevents error accumulation, at the cost of a naive discretization. CTPG (Sec.~\ref{sec:ppg}) discretizes adaptively, and stores a spline approximation of the trajectory to avoid error accumulation in the adjoint.
    }
    \label{fig:illustration}
    \vspace{-7mm}
\end{figure}

Adjoint sensitivity analysis~\citep{pontryagin1962mathematical} characterizes $\frac{\partial \mathcal{L}(x_0,\theta)}{\partial\theta}$ in a form that is amenable to computational approximation. Consider the adjoint process $\adjoint(t) \triangleq \frac{\partial \mathcal{L}(x_0,\theta)}{\partial x(t)}$; the evolution of this adjoint process for a trajectory $x$ can be described by the dynamics
\begin{equation}\label{eqn:adjoint}
    \frac{d\adjoint(t)}{dt} = -\adjoint(t)^\top \frac{\partial f}{\partial x(t)}-\frac{d w(x(t),\pi_\theta(x(t)))}{d x(t)}.
\end{equation}
These dynamics can be compared to \cite{chen2018neural}, with the notable distinction that the loss in Eq. \eqref{eqn:trajectoryloss} is path-dependent. This causes the instantaneous loss $w(x(t),u(t))$ to accumulate continuously in the adjoint process $\alpha(t)$. The adjoint process itself is constrained by the final value $\adjoint(T) = \frac{\partial J}{\partial x(T)}$. Sensitivity of the loss to the parameters $\theta$ is given by
\begin{equation}\label{eqn:gradloss}
\frac{\partial \mathcal{L}(x_0,\theta)}{\partial\theta} = \int_0^T \alpha(t)^\top \frac{\partial f}{\partial u} \frac{\partial u}{\partial \theta} + \frac{\partial w}{\partial u} \frac{\partial \pi_\theta(x(t))}{\partial \theta}\,dt.
\end{equation}
The integral in Eq. \eqref{eqn:gradloss} is amenable to estimation using various numerical techniques, each of which is susceptible to different forms of numerical error; a stylized illustration of the behaviour of various gradient estimators is presented in Figure \ref{fig:illustration}. 

\section{Numerical Policy Gradient Estimators}\label{sec:algos}

Policy gradients can be calculated via the integral described in Eq. \eqref{eqn:gradloss}. This integral in turn depends upon the trajectory $x(t)$ and the adjoint process $\alpha(t)$. To estimate a policy gradient, we require solutions to the following three sub-problems:
\begin{enumerate}
    \item The trajectory $x(t)$, governed by the dynamics in Eq. \eqref{eqn:dynamics}.
    \item The adjoint process $\alpha(t)$, governed by the dynamics in Eq. \eqref{eqn:adjoint}.
    \item The policy gradient accrued along $x(t)$ and $\alpha(t)$, described by the integral in Eq. \eqref{eqn:gradloss}. 
\end{enumerate}

The accumulated $\theta$ gradients, Eq. \eqref{eqn:gradloss}, can be described as the value at $t = 0$ of the process $g(t)$, with final value $g(T)= 0$, defined by the dynamics $\frac{dg(t)}{dt} = -\alpha(t)^\top \frac{\partial f}{\partial u} \frac{\partial u}{\partial \theta} - \frac{\partial w}{\partial u} \frac{\partial \pi_\theta(x(t))}{\partial \theta}$. 
Collectively, the trajectory process $x(t)$, the corresponding adjoint process $\adjoint(t)$, and the accumulated loss $g(t)$ comprise the following two-point boundary value problem (BVP):
\vspace{1mm}
\begin{align}
&\text{Initial Value}  & &\text{Dynamics} & & \text{Final Value}\notag\\
&x(0) = x_0,  & \frac{dx(t)}{dt} &= {\color{cornellred}f(x(t),u(t))},\label{eqn:forward}\\
&& \frac{d\adjoint(t)}{dt} &= - \adjoint(t)^\top {\color{cornellred}\frac{\partial f}{\partial x}}-{\color{darkcerulean}\frac{d w}{d x(t)}}, & & \alpha(T) = {\color{darkcerulean}\frac{dJ}{dx(T)}},\label{eqn:adjointfvp}\\
&& \frac{dg(t)}{dt} &= -\alpha(t)^\top {\color{cornellred}\frac{\partial f}{\partial u}}{\color{darkcerulean}\frac{\partial\pi_\theta}{\partial\theta}} - {\color{darkcerulean}\frac{\partial w}{\partial u}} {\color{darkcerulean}\frac{\partial \pi_\theta}{\partial \theta}}, & & g(T) = 0.\label{eqn:backward}
\end{align}
To understand how various subsystems of the policy gradient estimator interact, we highlight quantities computed by the physics simulator in {\color{cornellred}red} and quantities computed by automatic (or symbolic) differentiation in {\color{darkcerulean}blue}. 

The most common approach to estimating $g(0) = \frac{\partial \mathcal{L}(x_0,\theta)}{\partial \theta}$ in the controls community is backpropagation through time (BPTT). The idea is to discretize the dynamics $x(t)$ with a constant time step $h$, after which sensitivities to the policy parameters $\theta$ are computed by the standard backpropagation algorithm. In Section \ref{sec:bptt} we describe how this algorithm can be interpreted as an approximate solution to the continuous boundary-value problem, using Euler integration to estimate the dynamics $x(t)$, $\adjoint(t)$, and $g(t)$. The numerical ODE community widely eschews Euler integration in favor of more sophisticated solvers.
A natural question arises: can we improve upon naive Euler integration to construct better gradient estimates?

In Section \ref{sec:ppg} we introduce a new, general class of gradient estimators for this boundary value problem that we call the Continuous-Time Policy Gradient. In Section \ref{sec:bptt} we show how BPTT can be viewed as an instantiation of Continuous-Time Policy Gradients, using Euler integration as a numerical solver. The CTPG generalization allows us to replace Euler integration with a more sophisticated solver; in our experiments we opted for Runge-Kutta (RK4) and adaptive Adams-Moulton methods but these solvers can be seamlessly replaced by whatever solver is most suitable for a particular problem. In Section \ref{sec:neuralode} we discuss the Neural ODE estimator and its inadequacy for use with stabilizing control problems. And in Appendix \ref{sec:bvp} we discuss alternate methods for solving the BVP that may be of interest in specialized settings.

\subsection{Continuous-Time Policy Gradients}\label{sec:ppg}
We propose to solve the boundary value problem described by Eqs. \eqref{eqn:forward}, \eqref{eqn:adjointfvp}, and \eqref{eqn:backward} using a forward-backward meta-algorithm, which is analogous to (but distinct from) the forward and backward passes of an automatic differentiation system (and more generally, the forward-backward approach to dynamic programming). First, we use a numerical solver to estimate the solution to the initial value problem given by Eq. \eqref{eqn:forward} (the forward pass). Crucially we store the estimated trajectory for later use, incurring a storage cost off $O(s)$ where $s$ is the number of steps visited by the numerical solver. Second, we use a numerical solver to estimate the solution to initial value problem given by Eq. \eqref{eqn:adjointfvp} (the backward pass) using the cached trajectory computed in the forward pass. As we compute the backward pass we accumulate an estimate of the integral in Eq. \eqref{eqn:backward}, the total sensitivity of the loss to $\theta$ along the estimated trajectory. Details are presented in Algorithm~\ref{alg:ppg}.

\begin{wrapfigure}[18]{r}{0.6\textwidth}
    \centering
    \includegraphics[width=0.6\textwidth]{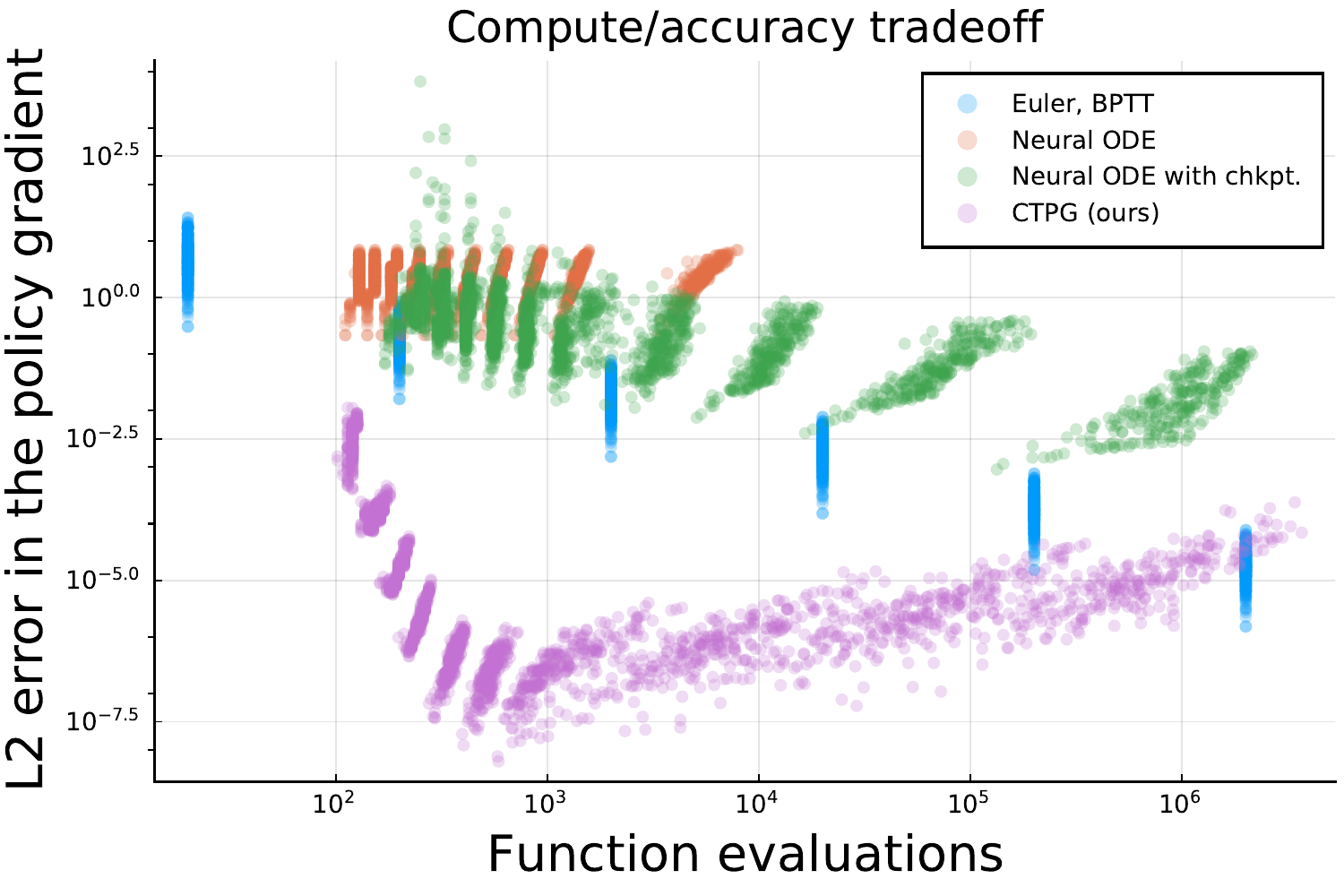}
    \caption{\vspace{-9mm}Approaching the Pareto frontier: Euler integration and other methods are Pareto dominated by CTPG. Lower and to the left is better. \vspace{5mm}}
    \label{fig:lqr_tradeoff}
\end{wrapfigure}

Figure \ref{fig:lqr_tradeoff} illustrates performance of Continuous-Time Policy Gradients (using the RK4 solver) and several other algorithms. The dynamics are given by the linear-quadratic regulator (LQR) and the policy is fixed to be the optimal LQR policy. By adjusting the error tolerance of the RK4 solver, we construct a trade-off curve between the accuracy of the CTPG estimates and the computational cost of computing an estimate. The CTPG estimator Pareto-dominates the other estimators.
In this sense, we argue that the other estimators are inadmissible: e.g. for any BPTT estimator, there is a CTPG estimator that is either (1) more accurate at the same level of performance or (2) more efficient at the same level of accuracy.

The effectiveness of CTPG is attributable to the efficiency of the RK4 solver, which can estimate the trajectory and adjoint paths while visiting many fewer states than a comparably accurate Euler solver (BPTT). This also means that CTPG is more memory efficient than BPTT: both algorithms require storage that grows linearly in the number of states visited in the forward pass. The (possibly surprising) under-performance of the Neural ODE estimator is explained in Section \ref{sec:neuralode}.

\begin{algorithm}
\SetAlgoLined
\vspace{1mm}
\SetKwInOut{Input}{Given}
\Input{Differentiable physics simulator $f(x, u)$, cost/reward function $w(x, u)$, and a numerical ODE solver $\text{Solve}[\text{initial\_conditions}, \text{dynamics}]$}
\SetKwInOut{Input}{Input}
\Input{Policy $\pi_\theta(x)$, initial state $x_0$}

\KwResult{An approximation of $\frac{\partial \mathcal{L}(x_0,\theta)}{\partial \theta}$}
\vspace{-1mm}\hrulefill\vspace{1mm}

\textbf{Forward pass:} (compute and store an approximation of the trajectory $x : [0,T] \to \mathbb{R}^d$)\\
\vspace{1mm}
\quad\quad $\tilde{x}(\cdot) \gets \text{Solve}\left[x(0) = x_0, \displaystyle\frac{dx(t)}{dt} = f(x, \pi_\theta(x))\right]$. 

\textbf{Backward pass:} (compute an approximation of the Pontryagin adjoints $\adjoint : [T,0] \to \mathbb{R}^d)$\\
\vspace{1mm}
\quad\quad $\tilde{\adjoint}(\cdot) \gets \text{Solve}\left[\adjoint(T) = \displaystyle\frac{d J}{d \tilde{x}(T)}, \frac{d \adjoint(t)}{d t} = - \adjoint(t)^\top \frac{\partial f}{\partial \tilde{x}(t)} - \frac{d w}{d \tilde{x}(t)} \right]$. 

\textbf{Return:}\\
\vspace{1mm}
\quad\quad $\displaystyle\int_0^T \tilde{\adjoint}(t)^\top \frac{\partial f}{\partial u} \frac{\partial u}{\partial \theta} + \frac{\partial w}{\partial u} \frac{\partial \pi_\theta}{\partial \theta}\,dt$\hspace{1mm} as an approximation of $\displaystyle\frac{\partial \mathcal{L}(x_0,\theta)}{\partial\theta}$.


\caption{Continuous-Time Policy Gradients}
\label{alg:ppg}
\end{algorithm}



\vspace{-2mm}
\subsection{Backpropagation Through Time}\label{sec:bptt}
\vspace{-.5mm}

Backpropagation through time (BPTT) is simply the application of reverse-mode automatic differentiation (AD) to computations involving a ``time'' axis. Like CTPG, AD proceeds with a forward and backward pass. In the forward pass, the trajectory $x(t)$ is approximated using discrete, fixed-length linearized steps (see Figure \ref{fig:illustration}); this is equivalent to using the Euler numerical integration method to solve the initial value problem Eq. \eqref{eqn:forward}. States computed in the forward pass are stored, incurring an $O(1/h)$ memory cost, and avoiding the need to recompute these values in the backward pass. In the backward pass of AD computes adjoints of the forward computation graph, using the values $x(t)$ stored during the forward pass. This is equivalent to using Euler integration to solve the final value problems described by Eqs. \eqref{eqn:adjointfvp} and \eqref{eqn:backward}. 

Backpropagation is an exact algorithm, in the sense that it computes exact gradients of the loss for the discretized dynamics. But these dynamics are only an approximation of the continuous trajectory $x(\cdot)$, and therefore the gradients only approximate the desired value $g(0)$. Computing an accurate estimate of the continuous trajectory using Euler integration requires an excessively small step size $h$, and a correspondingly large memory allocation. Furthermore, the resulting computation graph is very deep, which could make learning difficult \citep{quaglino2019snode}.

\subsection{The Neural ODE Algorithm}\label{sec:neuralode}

\begin{figure}[h!]
    \centering
    \includegraphics[width=0.49\textwidth]{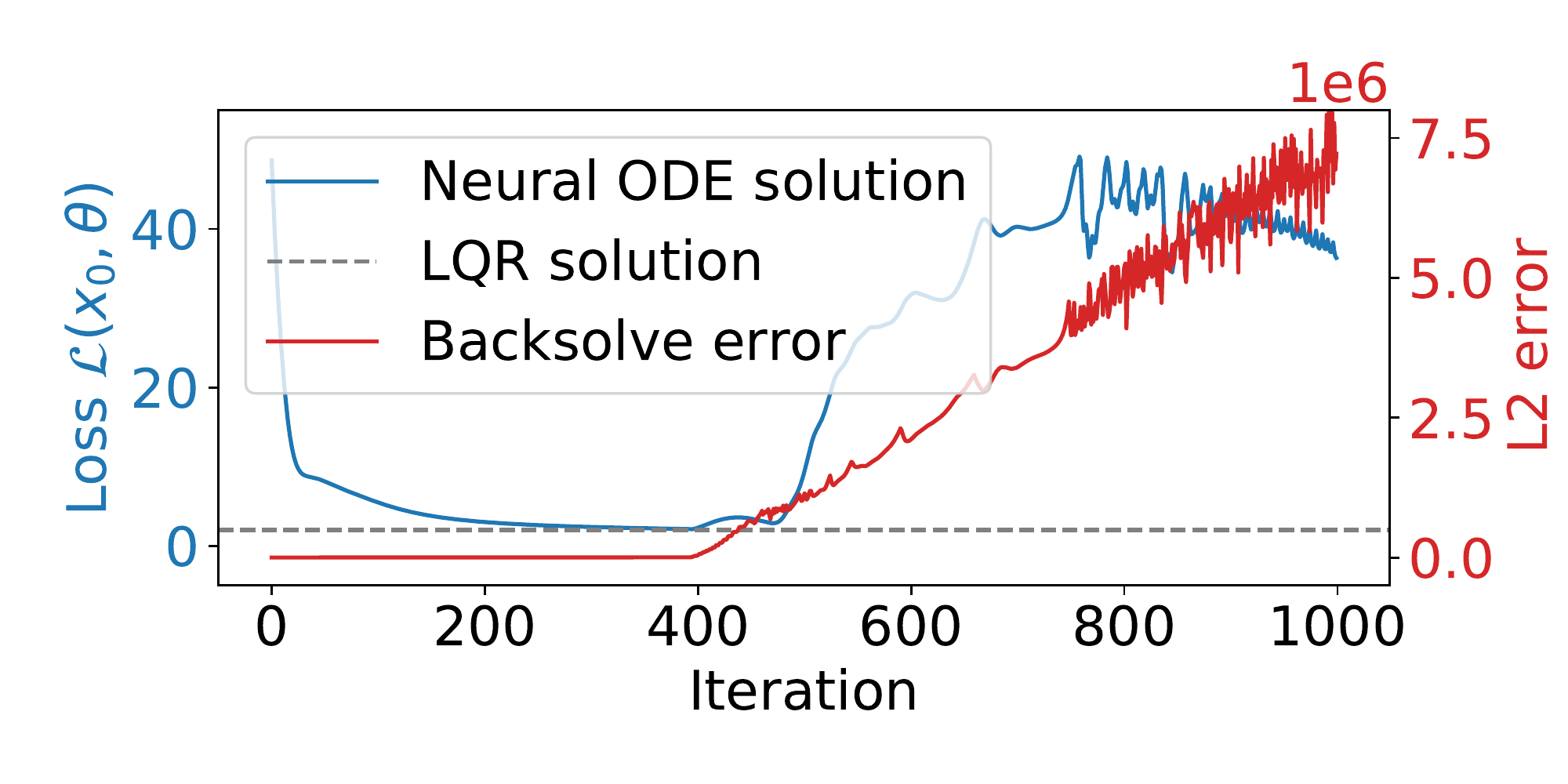}
    \includegraphics[width=0.49\textwidth]{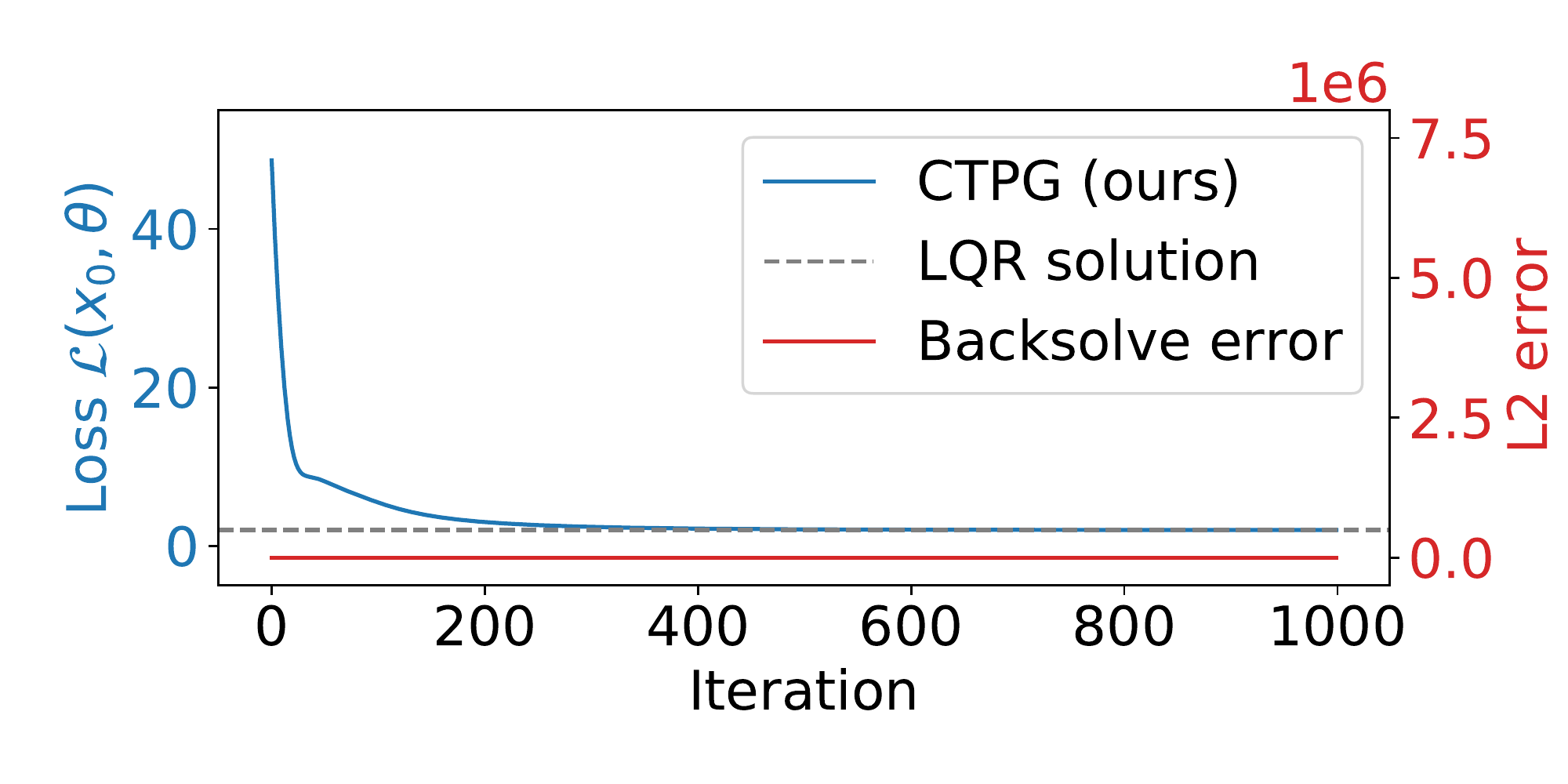}
    \caption{\vspace{-5mm}Left: The Neural ODE estimator used to learn an LQR policy. The trajectory loss $\mathcal{L}(x_0, \theta)$ is shown in blue, and the discrepancy
    $||x(0) - \tilde{x}(0)||^2$ is shown in red where $\tilde{x}(t)$ is the ``recovered'' path taken by the adjoint process. The discrepancy between the forward and reverse paths is negligible until the control learns to stabilize the system, at which point it becomes impossible to recover $x(0)$ from $x(T)$. Right: The same problem, but with CTPG. Unlike the Neural ODE, CTPG records a lightweight spline estimate of $x(t)$.
    \vspace{-8mm}}
    \label{fig:linear_control_example}
\end{figure}

To efficiently compute the backward pass, we rely on a stored spline of states $s$ visited from the forward pass, leading to an algorithm with $O(s)$ space complexity. For many physical systems, the dynamics $f(x,u)$ are ``invertible''; this invites us to consider algorithms with constant space complexity, recomputing the state variables $x(t)$ in the backward pass following the reverse dynamics $-f(x,u)$ \citep{chen2018neural}. Constant-memory algorithms are intriguing, but we observe that their results are numerically unstable for stabilizing control problems. We visualize this instability for an LQR system in Figure \ref{fig:linear_control_example}.



In principle, a trajectory $x(t)$ is uniquely defined by either its initial value $x(0)$ thanks to the Picard-Lindelöf theorem. 
But for control problems that seek to stabilize the system towards some fixed goal-point, it is numerically unstable to reconstruct the trajectory from its final state; moreover, this instability becomes more pronounced as the policy learns to more effectively achieve its goal, and computing the inverse trajectory becomes increasingly chaotic. 

To better formalize this notion, we appeal to linear stability theory. For a system $\frac{dx}{dt} = \phi(x)$, linear stability analysis suggests inspecting the eigenspectrum of $\frac{d\phi}{dx}$: if $\frac{d\phi}{dx}$ has all eigenvalues with negative real part the system is said to be linearly stable, and if there exists an eigenvalue with positive real part it is said to be linearly unstable. Stabilizing control problems by their nature demand solutions such that the system has negative eigenvalues $-\lambda \ll 0$, as small as possible subject to control costs. This becomes catastrophic in the reverse-direction when we are faced with a linearly unstable system with eigenvalues $\lambda \gg 0$! Furthermore, as we discuss in Appendix~\ref{sec:instability}, the neural ODE backpropagation dynamics are in fact linearly unstable everywhere.


\section{Experiments}\label{sec:exp}

To better understand CTPG's behavior, we compare CTPG with BPTT for a variety of control problems. We restrict our experiments to settings in which first-order policy gradients optimization can effectively discover the optimal policy; extension of these local search algorithms to global policy optimization is beyond the scope of this work. We begin with a relatively straightforward control problem in Sec.~\ref{sec:diffdrive} for which we wrote a simple differentiable simulator. We then evaluate CTPG for a task defined in an existing differentiable physics engine from \cite{hu2019difftaichi} in Sec.~\ref{sec:electric}. In Sec.~\ref{sec:cartpole} we evaluate CTPG for a task defined in the MuJoCo physics simulator, using finite difference approximations to the gradients of the physical system. We conclude in Sec.~\ref{sec:quadrotor} with a challenging quadrotor control problem running in a purpose-built differentiable flight simulator. Code to reproduce our experiments is available at \url{https://github.com/samuela/ctpg}.

\subsection{Differential Drive Robot}\label{sec:diffdrive}
\begin{figure}[h]
    \centering
    \includegraphics[trim=50 0 50 0, clip,width=0.32\textwidth]{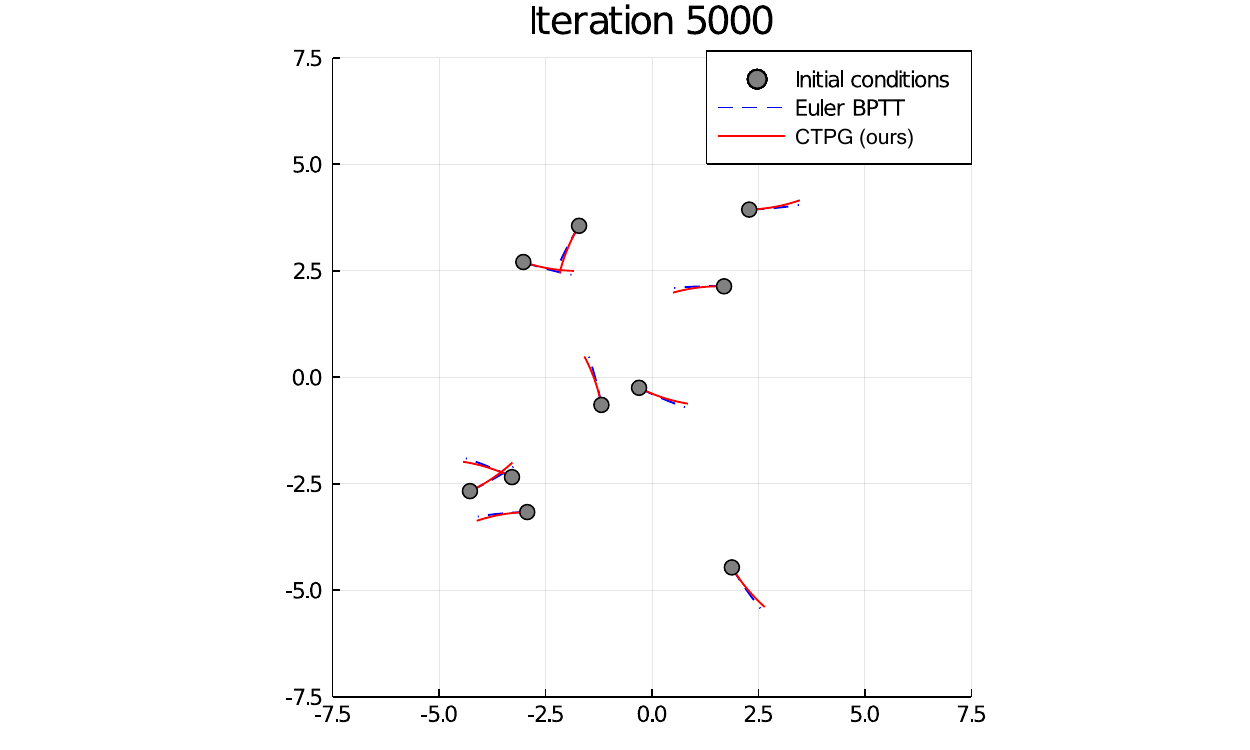}
    \includegraphics[trim=50 0 50 0, clip,width=0.32\textwidth]{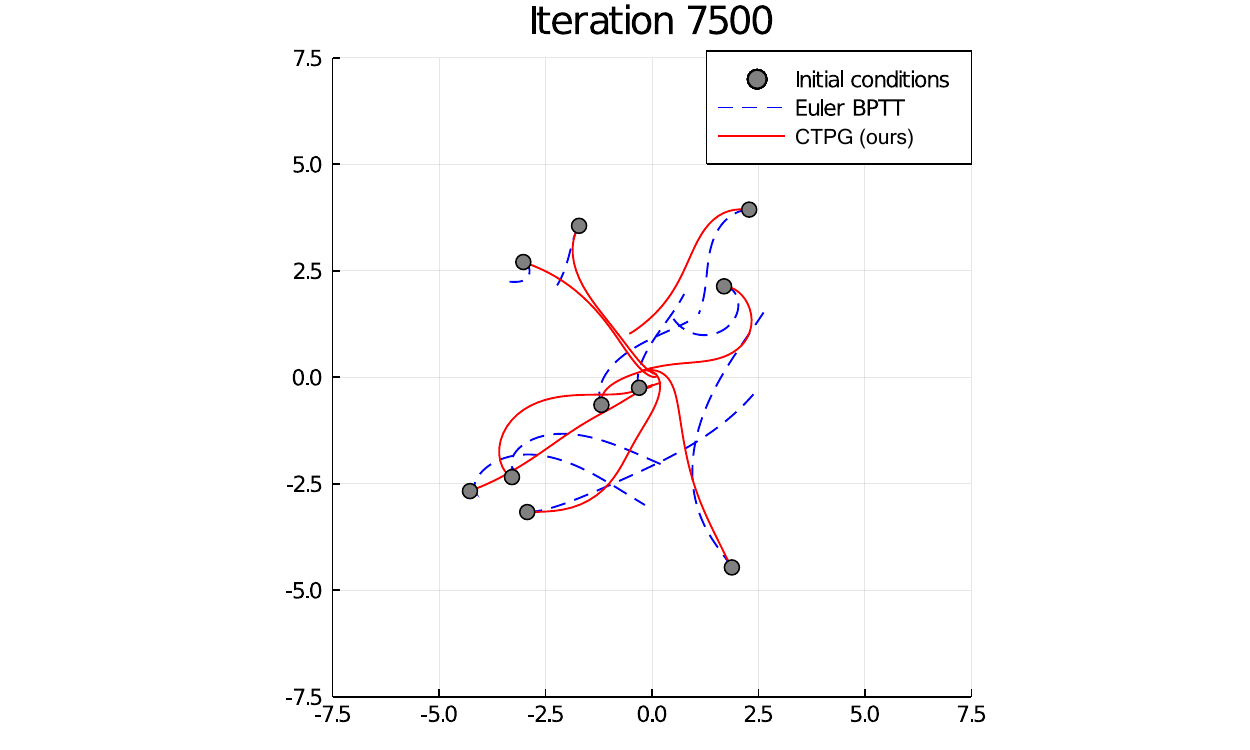}
    \includegraphics[trim=50 0 50 0, clip,width=0.32\textwidth]{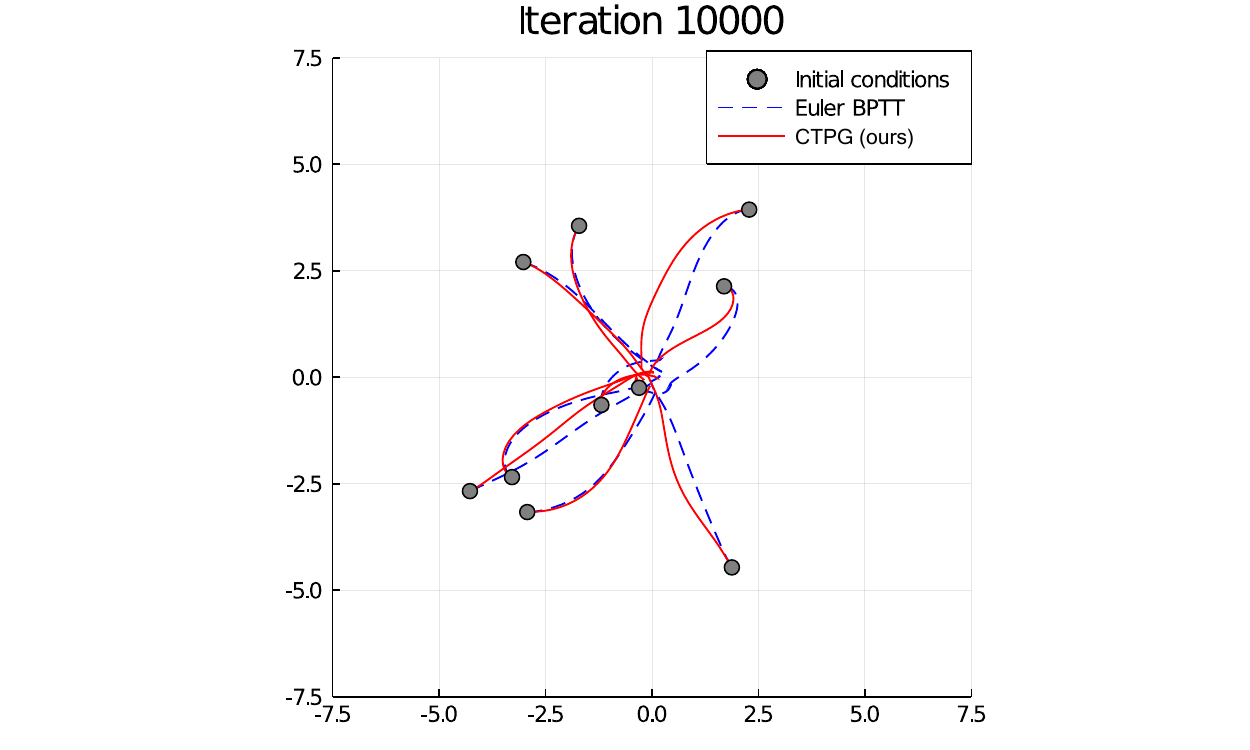}
    \caption{Stages of training policies to control a differential drive robot to stabilize towards the origin. Each curve denotes the trajectory of the robot in the $(x,y)$ plane under the respective policies. A full video is available in the supplementary material. 
    }
    \label{fig:diffdrive_ppg_vs_euler}
\end{figure}

The differential drive is a ``Roomba-like'' robot with two wheels that are individually actuated to rotate or advance the robot. We use the dynamics defined in \cite{lavalle2006planning}
, but control the torque applied to each wheel rather than directly controlling their angular velocities. We initialized the robot to random positions and rotations throughout the plane and learn a policy that drives it to the origin. See Figure~\ref{fig:diffdrive_ppg_vs_euler} for a visualization of results. Both CTPG and BPTT are able to learn working policies, but CTPG is able to do so more efficiently. Videos of the training process are available in the supplementary material.

\subsection{DiffTaichi Electric Field Control}\label{sec:electric}

To test CTPG's ability to work with third-party differentiable physics simulators, we integrated and compared against the DiffTaichi (meta) physics engine \citep{hu2019difftaichi, hu2019taichi}. DiffTaichi (now part of Taichi) is a differentiable, domain-specific language for writing physics simulators. We tested against their electric field control experiment in which eight fixed electrodes placed in a 2-D square modulate their charge driving a red ball with a static charge around the square. The red ball experiences electrostatic forces given by Coulomb's law, $\mathbf{F} = k_e \frac{q_1 q_2}{|\mathbf{r}|^2} \hat{\mathbf{r}}$ \citep{coulomblaw}.

\begin{figure}[h]
    \centering
    \includegraphics[height=1.3in]{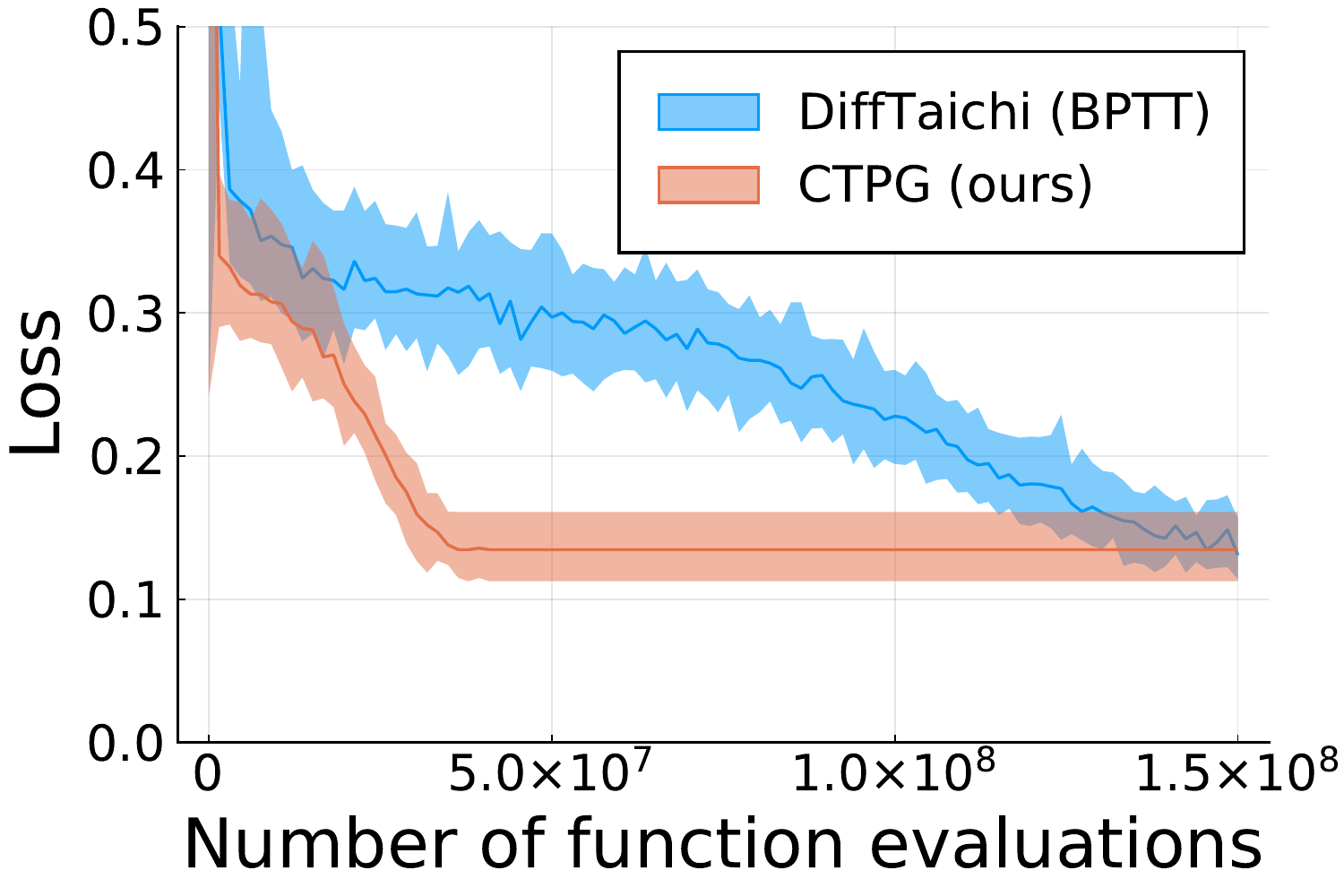}
    \includegraphics[height=1.3in]{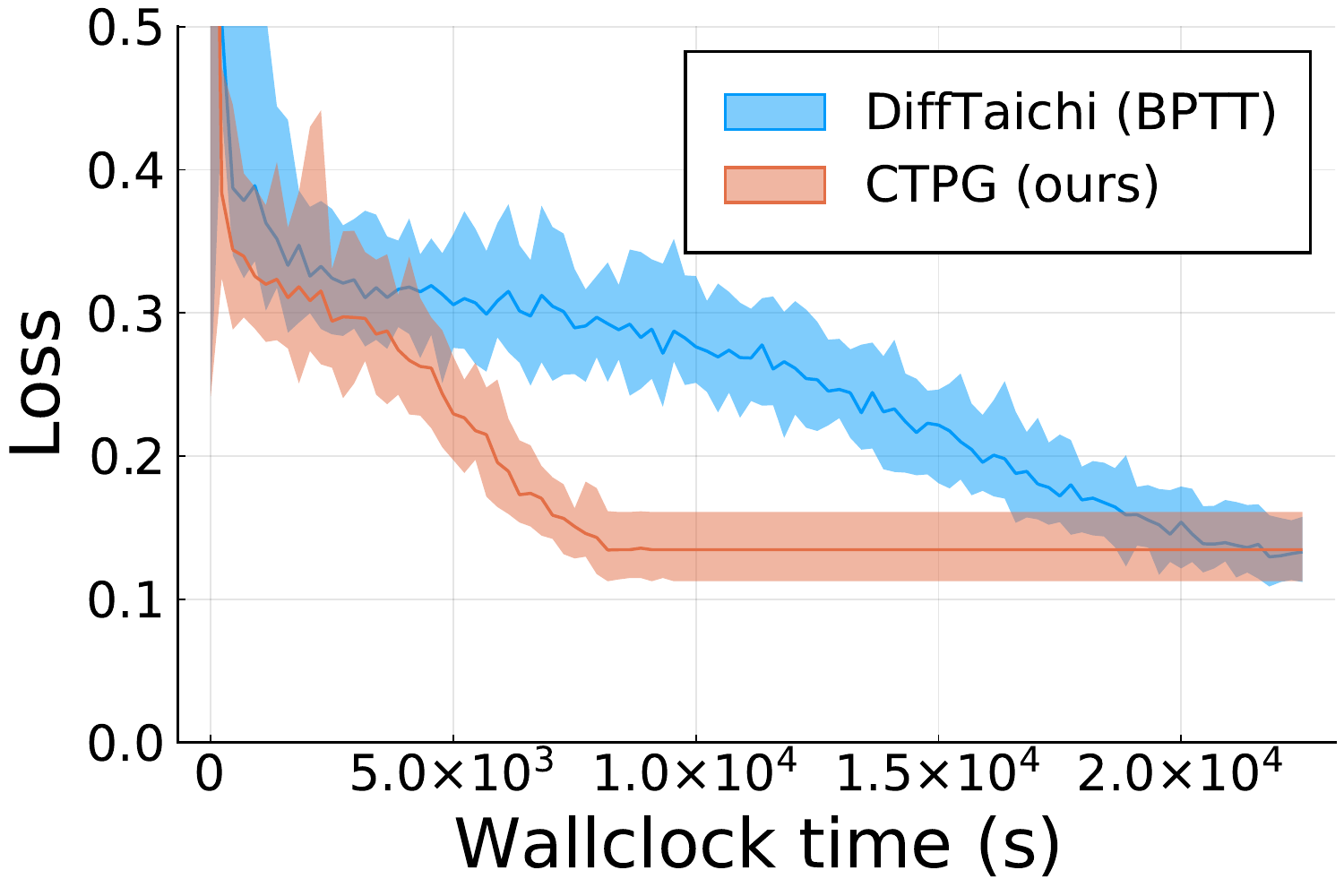}
    \hspace{0.15in}
    \includegraphics[height=1.3in]{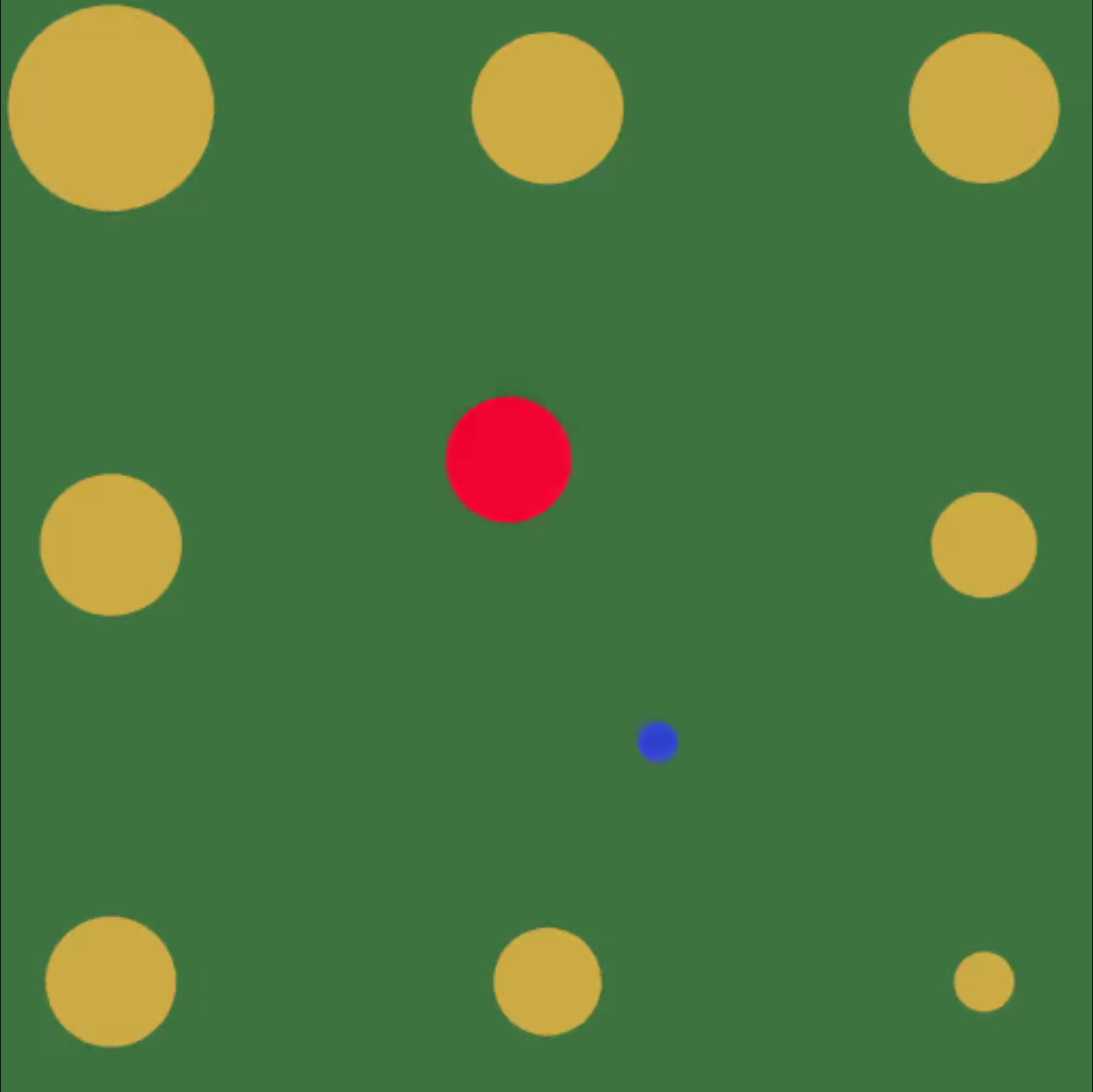}

    \caption{The DiffTaichi electric experiment involves modulating the charges of the 8 yellow electrodes to push the red charge to follow the blue dot (right). We found that CTPG performance was dominant in terms of the number of function evaluations (left) as well as wallclock time (center) across 32 random trials.}
    \label{fig:electric_ppg_vs_bptt}
\end{figure}

We evaluate performance both in terms of wallclock time and in terms of the number of oracle function calls made to the simulator, namely the combined evaluations of $f$, $\partial f/ \partial x$, and $\partial f / \partial u$. Calls to the simulator constitute the vast majority of the run time, so we find this to be the most accurate hardware-independent measure of performance. We present the results in Fig.~\ref{fig:electric_ppg_vs_bptt}. We find that CTPG reliably outperforms BPTT both in terms of wallclock time and number of function evaluations, even with DiffTaichi's exact BPTT configuration.

\subsection{MuJoCo Cartpole}\label{sec:cartpole}

MuJoCo \citep{Todorov2012MuJoCoAP} is a widely used physics simulator in both the robotics and reinforcement learning communities. In contrast to DiffTaichi (Section \ref{sec:electric}) MuJoCo models are not automatically differentiable, but rather they are designed to be finite differenced. We demonstrate that BPTT and CTPG can be applied to such black-box physics engines. These finite difference derivatives are used in Eqs. \ref{eqn:adjointfvp} and \ref{eqn:backward} alongside the automatically differentiated policy derivatives. We test on the Cartpole swing-up task specified by the dm\_control suite \citep{tassa2020dmcontrol}, a classic non-linear control problem. This model was re-implemented in the Lyceum framework \citep{lyceum} with the Euler integration timestep used for BPTT kept the same as the model specification (0.01 seconds). Total dynamics evaluations included those used for finite differencing.

\begin{figure}[!t]
    \centering

    \raisebox{8mm}{\includegraphics[trim=300 0 150 90,clip,width=.25\textwidth]{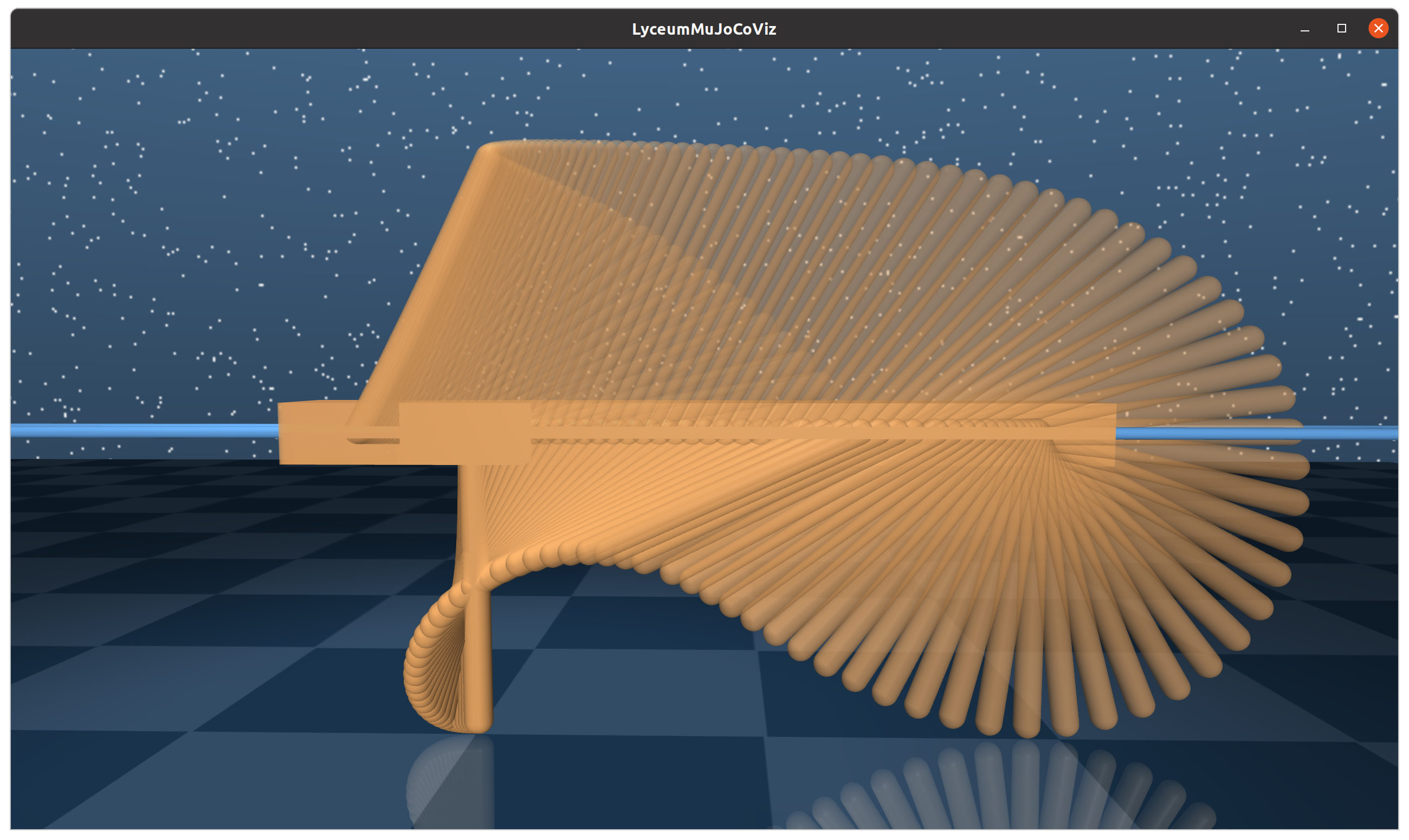}}
    \raisebox{8mm}{\includegraphics[trim=300 0 150 90,clip,width=.25\textwidth]{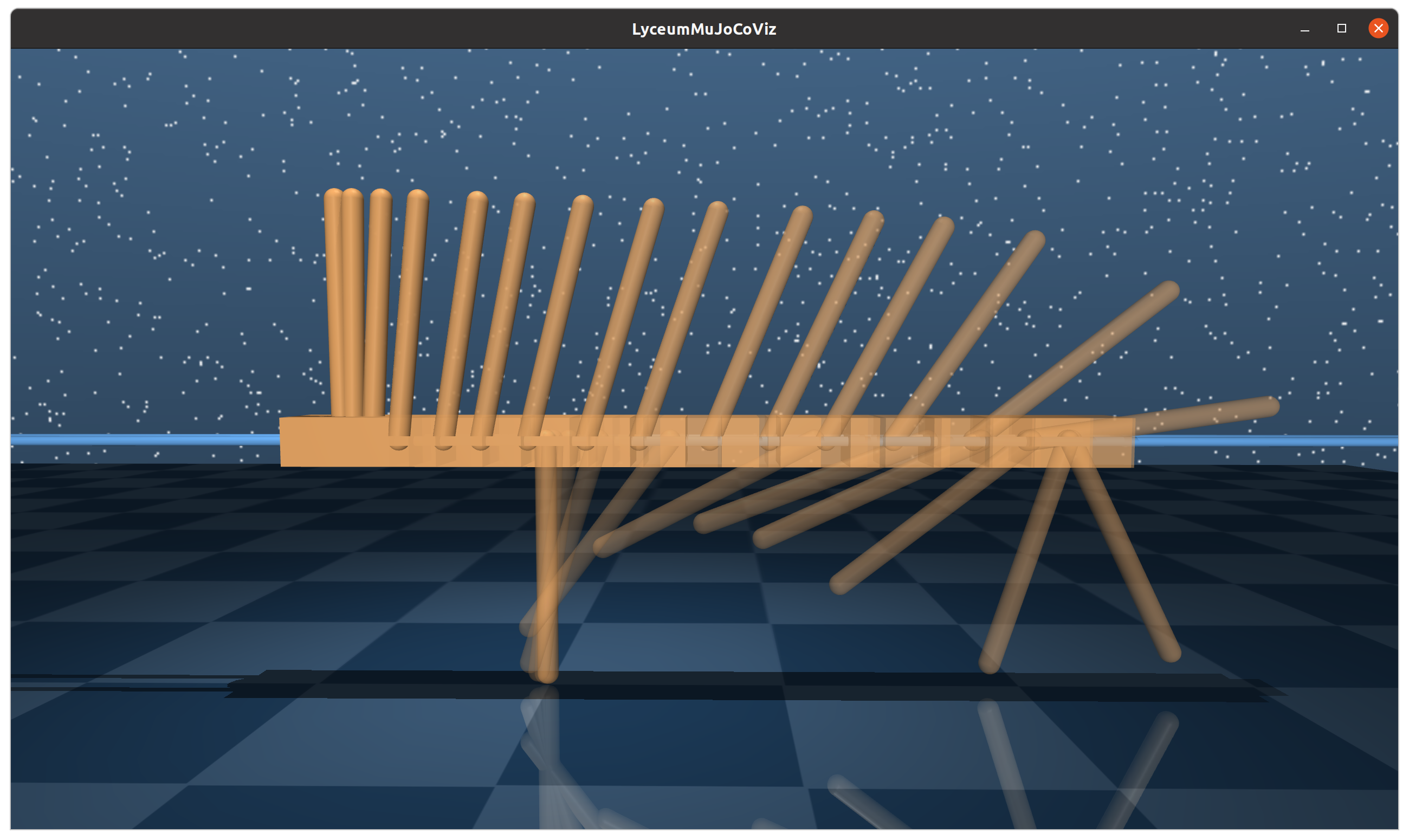}}
    \hspace{-3mm}
    \includegraphics[width=.4\textwidth]{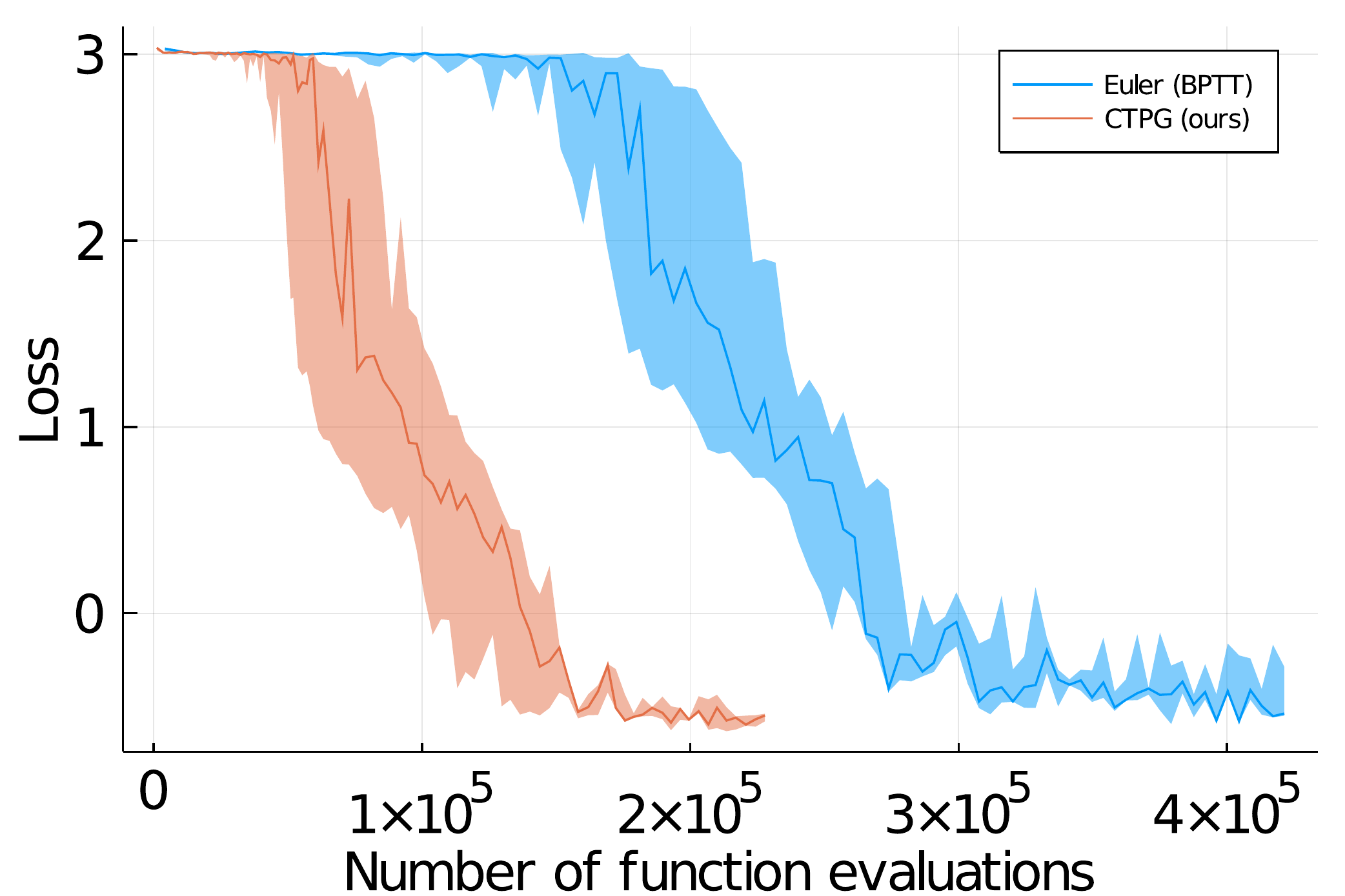}
    \vspace{-2mm}
    
    
    \caption{Visualization of all states visited by a policy trained with BPTT (left) and CTPG (middle). Fewer states are visited by CTPG during training leading to more efficient learning (right). Error bars on the training plot show the 0.25-0.75 quantile range of 5 random seeds. 
    \vspace{-5mm}}
    \label{fig:cartpole_ctpg_vs_euler}
\end{figure}

Results are presented in Fig.~\ref{fig:cartpole_ctpg_vs_euler}. We find that CTPG needs significantly fewer dynamics evaluations than BPTT per learning instance. While both techniques eventually reach a similar level of performance for this task, CTPG demonstrates substantial efficiency gains. 

\subsection{Quadrotor Control}\label{sec:quadrotor}

\begin{wrapfigure}[15]{r}{0.6\textwidth}
    \vspace{-5mm}
    \centering
    \includegraphics[width=0.5\textwidth]{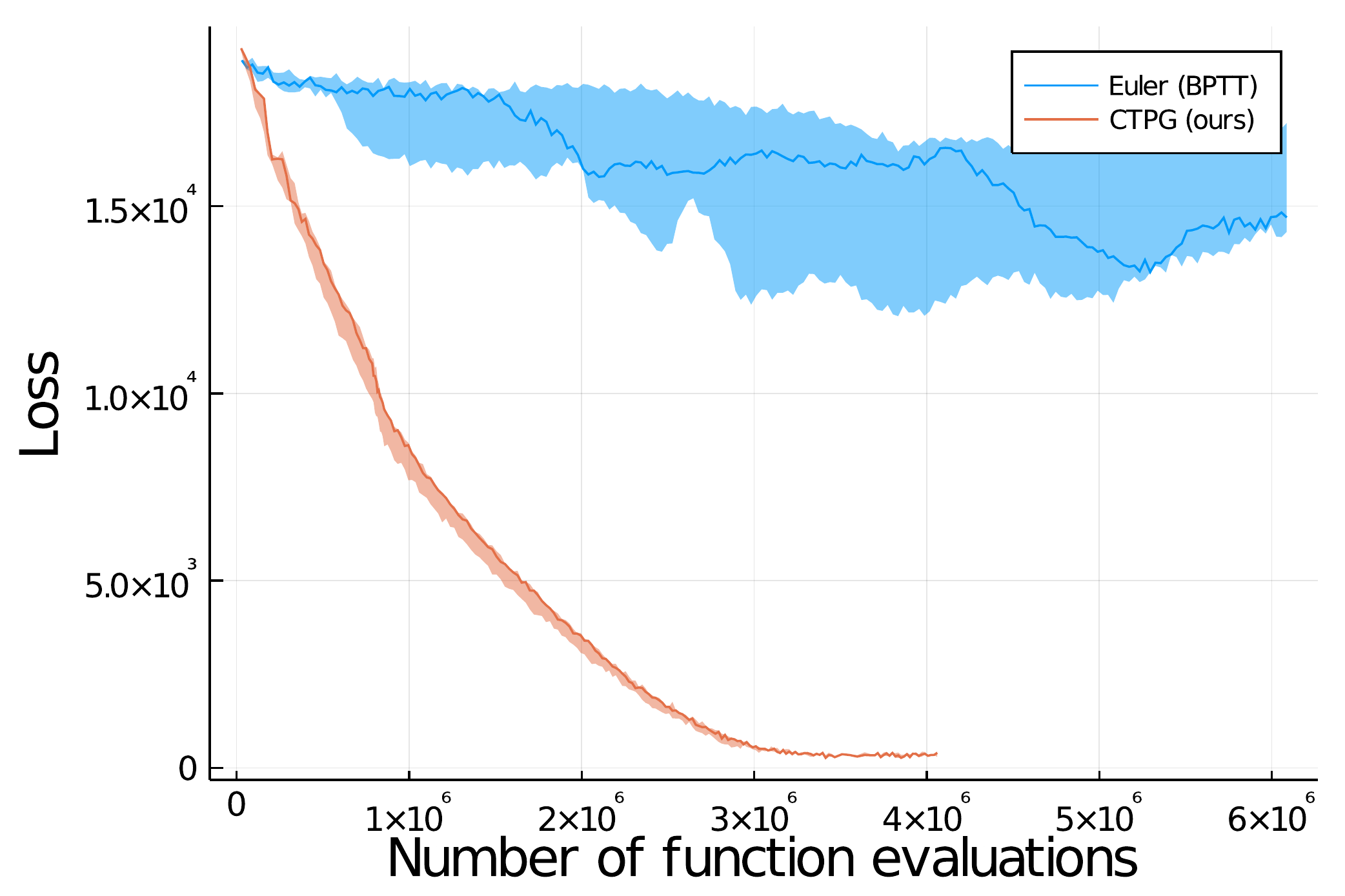}
    \vspace{-5mm}
    \caption{
    CTPG learns policies to stabilize the quadrotor towards the origin, while Euler integration and BPTT--with the same number of function evaluations--can not. 
    }
    \label{fig:quadrotor_ppg_vs_euler}
\end{wrapfigure}

Quadrotors are notoriously difficult to control, especially given the strict real-time compute requirements of a flying aircraft, and tradeoffs between compute power, weight, and power draw. As such, they make for a strong use case of the policy learning tools that CTPG brings to bear. We test CTPG's ability to learn a small, efficient neural network controller that flies a quadrotor. 

We implemented a differentiable quadrotor simulator based on the dynamics from \cite{Sabatino2015QuadrotorCM}, and train policies to stabilize a quadrotor at the origin from a wide distribution of initial positions, orientations, and velocities. Our experimental setup is similar to that of \cite{8593536}. Results are presented in Fig.~\ref{fig:quadrotor_ppg_vs_euler}.
As a flying robot, the quadrotor dynamics are inherently unstable. As a result, we found that Euler integration is often not able to integrate trajectories to sufficient accuracy for policy learning using a similar sample budget to CTPG. 

\section{Related Work}\label{sec:rel}


\textbf{Neural ODEs.} \citet{chen2018neural} introduced neural ordinary differential equations, based on the adjoint sensitivity analysis method of \citet{pontryagin1962mathematical} and using backwards dynamics to reconstruct $x(t)$ as described in Eq.~\ref{eqn:forward}. Recent work by \cite{Du2020ModelbasedRL} and \cite{quaglino2019snode} studied using Neural ODEs for system identification as opposed to policy learning. \cite{ANODE} explored a gradient checkpointing scheme for Neural ODEs.

\textbf{Differentiable physics.} Differentiable physics engines have recently attracted attention as a richer alternative to conventional ``black-box'' simulators. \cite{e2ediffphysics_NEURIPS2018_842424a1} introduced a differentiable rigid-body physics simulator based on analytic derivatives through physics defined via a linear complementarity problem. \cite{Qiao2020Scalable} expanded on the types of physical interactions that could be efficiently simulated and differentiated.~\citet{li2020learning, li2018propagation} and~\citet{Battaglia2016InteractionNF} proposed developing differentiable physics engines by learning them from data. \cite{Schenck-RSS-17,pmlr-v87-schenck18a} built a differentiable simulator for liquids with applications to robotics. \cite{hu2019taichi,hu2019difftaichi} created a differentiable domain-specific language for the creation of physics simulators, enabling backpropagation through a variety of different physical phenomena.

\textbf{Trajectory optimization and control theory.} \cite{pontryagin1962mathematical} introduced necessary conditions for optimality of a continuous trajectory, drawing from the calculus of variations. An alternative perspective, the Hamilton-Jacobi-Bellman equation, provides necessary and sufficient conditions for optimality of policies~\citep{hjb}. Prior work describes the analytic adjoints for backpropagation through time \citep{underactuatedTedrake,robinson:fallside:nips:1988,werbos_paul_j_1988_1258627,mozer1995}. Unlike the majority of work in trajectory optimization and control, we use Pontryagin's maximum principle to learn global feedback policies instead of fitting single trajectories. Concurrently \cite{jin2019pontryagin} explored differentiating through Pontryagin's maximum principle. However, they still considered standard time discretizations, eg. BPTT. Other works have investigated similar forward-backward schemes for solving the adjoint equations \citep{McAsey2012ConvergenceOT}.

\textbf{Numerical methods for ODEs.} The study of numerical methods for solving ODEs has a long and rich history, dating back to Euler's method \citep{solvingODEsbook}. In this paper we leveraged higher-order explicit solvers with adaptive step sizes, especially Runge-Kutta \citep{rg,tsitouras2011runge} and Adams-Moulton methods~\citep{depac,rl}. For a thorough treatment of ODE solvers we refer the reader to~\cite{solvingODEsbook}.
\section{Discussion}\label{sec:conclude}

We introduced Continuous-Time Policy Gradients (CTPG), a new class of policy gradient estimator for continuous-time systems. By leveraging advanced integrators developed by the numerical ODE community, CTPG enjoys superior performance to the standard back-propagation through time (BPTT) estimator. Although we studied policy gradients in this work, it's worth noting that CTPG is general enough to operate as a ``layer'' within any larger differentiable model. This work could be extended to other related scenarios: systems with non-deterministic dynamics (using stochastic ODE solvers) infinite-time planning problems (using steady state ODE solvers) and sensor/actuation latency (using delay differential equations solvers). Adjoints for these alternatives have already been implemented in \cite{rackauckas2017differentialequations}. 
Another compelling direction for future work would be the incorporation of value function learning via extension to the Hamilton-Jacobi-Bellman equation.
We are interested in the following questions: (1) When and why do policy gradient optimizations fail? 
(2) How can we properly incorporate policy gradient estimates into a global policy search?


\clearpage

\acks{Thanks to Behçet Açikmeşe, Byron Boots, Vincent Roulet, Sharad Vikram, Andrew Wagenmaker, and Matt Johnson for helpful discussions, support, and feedback. Thanks to Amanda Baughan for assistance with visualizations and feedback. Special thanks to Chris Rackauckas for his development and support of the SciML software suite. This work was (partially) funded by the National Science Foundation IIS (\#2007011), National Science Foundation DMS (\#1839371), the Office of Naval Research, US Army Research Laboratory CCDC, Amazon, and Honda Research Institute USA.}

\bibliography{references.bib}

\clearpage
\appendix

\section{Experimental models}

For complete experimental details, please see our code at~\url{https://github.com/samuela/ctpg}.

\subsection{Linear quadratic regulator}
As usual,
\begin{align}
w(x,u) &= x^\top Q x + u^\top R u,\\
f(x,u) &= Ax + Bu. 
\end{align}
For the example in Fig.~\ref{fig:linear_control_example}, we used $A = 0$, $B = Q = R = I$, $x_0 = [1, 1]^\top$ and $T=25$. The control policy was given by a single hidden layer neural network with 32 hidden units and tanh activations. Training was run for 1,000 iterations.

\subsection{Differential drive robot}
Following \url{http://planning.cs.uiuc.edu/node659.html}, the physical dynamics for a differential drive robot with left wheel speed $\omega_\ell$, right wheel speed $\omega_r$, position $x,y$, heading $\theta$, and wheelbase $L$ are given by,
\begin{align}
\frac{dx}{dt} = \frac{\omega_\ell+\omega_r}{2} \cos \theta, \quad \frac{dy}{dt} = \frac{\omega_\ell+\omega_r}{2} \sin \theta, \quad \frac{d\theta}{dt} = \frac{\omega_r - \omega_\ell}{L}.
\end{align}
Control is applied such that
\begin{equation}
\frac{d\omega_\ell}{dt} = u_l, \quad \frac{d\omega_r}{dt} = u_r.
\end{equation}
The instantaneous cost is defined as
\begin{align}
    w(x, y, \omega_\ell, \omega_r, u_l, u_r) = x^2 + y^2 + \frac{1}{10} (\omega_\ell^2 + \omega_r^2 + u_l^2 + u_r^2).
\end{align}
For the control policy we used a two hidden layer neural network with 64 units in each hidden layer and tanh activations. We additionally augment the input to the network with a few useful manual features including $\cos \theta$ and $\sin \theta$.







\subsection{DiffTaichi Electric}
We integrated with the DiffTaichi example code. This integration required refactoring the example code to return forces. For an apples-to-apples timing comparison we also wrote a wrapper BPTT implementation following exactly the same implementation as the reference code. We evaluated BPTT with the same dt, and other hyperparameters, as was specified in the DiffTaichi code. Experiments were run with 32 random trials, and plots show the 5\%-95\% percentile error bars.
Following \cite{hu2019difftaichi}, we used a single hidden layer neural network policy with 64 hidden units and tanh activations.

\subsection{MuJoCo Cartpole}

The specifications of the MuJoCo Cartpole experiment are as follows. MuJoCo derivatives were calculated through forward finite differencing (chosen over central for speed) with an epsilon of $1e-6$. Derivatives with respect to positions, velocities, and controls were all calculated in this manner. For reference, an epsilon of $1e-4$ to $1e-8$ all perform approximately the same. The policy was a two layer network with 32 hidden units and \texttt{tanh} activations. The last layer's initialization weights were scaled, in this case by 0.1, as is typical in reinforcement learning \cite{andrychowicz2020matters}. The policy function maps observations of the system to controls.

During optimization we used the ADAM optimizer with a step size of 0.001. The number of function evaluations we count includes those of the finite difference calculations. More specifically, every time the MuJoCo forward dynamics function \texttt{mj\_forward} was called was included. We use a mini-batch of 2 rollouts, with each starting state sampled according to the DMcontrol specifications; the sequence of starting states is set to be the same between BPTT and CTPG as well as the policy initializations. Both methods were run for a fixed 100 iterations, with the losses and number of function evaluations counted across 5 random seeds. 

\subsection{Quadrotor}

The quadrotor policy was a two layer network with 16 hidden units and \texttt{tanh} activations. The policy was trained with the ADAM optimizer with a step size of 0.01, with the gradient values clipped to $\pm 1.0$. Training was performed with 5 random seeds with a minibatch size of 32 per optimization step. The dynamics of the quadrotor experiment are given by \cite{Sabatino2015QuadrotorCM}.

\clearpage

\section{Neural ODEs and their linear stability properties}\label{sec:instability}

Consider the LQR control problem with dynamics $\frac{dx(t)}{dt} = Ax(t) + Bu(t)$. Under a linear feedback policy $u(t) = -K x(t)$, the system reduces to a first-order linear ODE: $\frac{dx(t)}{dt} = (A - BK)x(t)$. With any luck we will identify $K$ that \textit{stabilizes} the system so that $x(t)$ converges exponentially to $0$ everywhere. Consider running such a system forward to some time $T$, and then attempting to run it backward along the same path. Because all paths will converge to the origin, when starting at the origin and trying to run backwards it becomes numerically infeasible to recover the correct trajectory to $x(0)$.

\begin{claim*}\label{prop:node-linear-stability}
For any non-constant dynamics $f$, the dynamics of the Neural ODE backpropagation process will have eigenvalues $\pm\lambda_1, \dots, \pm\lambda_d$ split among the $x(t)$ and $\adjoint(t)$ process, along with zero eigenvalues for $g(t)$. Therefore the Neural ODE backpropagation process is linearly unstable for all $t$.
\end{claim*}
Let $\psi(t) = [x(t), \adjoint(t), g(t)]$ denote the combined Neural ODE backpropagation process. The Neural ODE stipulates that in the reverse time solve we begin with $\psi(T)$ and follow the dynamics $-\dot{\psi}(t)$ as given in Eqs.~\ref{eqn:forward}, \ref{eqn:adjointfvp}, and \ref{eqn:backward}. 
Denoting the eigenvalues of $\frac{\partial f}{\partial x}$ as $\lambda_1, \dots, \lambda_d$, we have
\begin{equation}
-\frac{\partial \dot{\psi}(t)}{\partial \psi(t)} = \begin{bmatrix}
-\frac{\partial f}{\partial x} & 0 & 0 \\
\adjoint(t)^\top \frac{\partial^2 f}{\partial x^2} & \left(\frac{\partial f}{\partial x}\right)^\top & 0\\
0 & \left(\frac{\partial f}{\partial \theta}\right)^\top & 0
\end{bmatrix}
\end{equation}
which has eigenvalues $\pm \lambda_1, \dots, \pm \lambda_d$, and 0 with multiplicity $n_\theta$. Note that a system is considered stable when $\text{Re}(\lambda_i) < 0$ for all such eigenvalues $\lambda_i$. However because all eigenvalues come in positive and negative pairs, we are doomed to experience eigenvalues with positive real parts, indicating eigen-directions of exponential blowup.


\clearpage
\section{Fusing Operations in the CTPG Algorithm}\label{sec:bvp}

One opportunity for computational efficiency that we take advantage of in our experiments is that $\adjoint(t)$ and $g(t)$ evolve in lockstep. This means that we do not need to store the estimates $\tilde \alpha(t)$ computed in Algorithm \ref{alg:ppg}, but instead accumulate $g(t)$ concurrently as we solve for $\tilde \adjoint(t)$. The most direct way to do this, while still treating the solver as a black box, is to simply concatenate or ``fuse'' the adjoint and gradient dynamics given by Equations \eqref{eqn:adjointfvp} and \eqref{eqn:backward} and perform a single backward solve to simultaneously compute the adjoints and accumulate the gradient. The automatic differentiation implementation of BPTT performs this fusion by default.

While the dynamics described by Eqs. \eqref{eqn:forward} and \eqref{eqn:adjointfvp} both evolve in $\mathbb{R}^d$, the dynamics \eqref{eqn:backward} evolve in the parameter space of the parameter space of the policy $\pi_\theta$, which can be large. Concatenating the dynamics on Eqs. \eqref{eqn:adjointfvp} and \eqref{eqn:backward} could  cause problem for numerical ODE solvers that are designed for low dimensional (e.g. physical) spaces. Nevertheless, we find that fusing these two computations works well for the policies and physics considered in this paper.

We could take this fusion perspective further and ask whether we can directly solve the full BVP, rather than sequentially computing the forward and backward passes. This is possible, and could be approached using a collocation algorithm. We favor the CTPG because, if our policy optimization is part of some larger robotic system that is sensitive to  downstream consequences of arriving at a final state $x(T)$, then we need an estimate $\tilde x(T)$ in order to define the boundary condition $\frac{dJ}{dx(T)}$. This means that CTPG could be used not just as a policy gradient estimator, but also as a layer in a larger end-to-end problem with a continuous-time submodule.

\end{document}